\documentclass[sigconf]{acmart}
\AtBeginDocument{%
  \providecommand\BibTeX{{%
    \normalfont B\kern-0.5em{\scshape i\kern-0.25em b}\kern-0.8em\TeX}}}

\copyrightyear{2023} 
\acmYear{2023} 
\setcopyright{acmlicensed}
\acmConference[MM '23]{Proceedings of the 31st ACM International Conference on Multimedia}{October 29-November 3, 2023}{Ottawa, ON, Canada}
\acmBooktitle{Proceedings of the 31st ACM International Conference on Multimedia (MM '23), October 29-November 3, 2023, Ottawa, ON, Canada}
\acmPrice{15.00}
\acmDOI{10.1145/3581783.3612281}
\acmISBN{979-8-4007-0108-5/23/10}

\usepackage{algorithm}
\usepackage{algorithmic}
\usepackage{subfigure}
\usepackage{multirow}
\usepackage{makecell}
\newcommand{\ie}{\textit{i}.\textit{e}.}

\newcommand{\eg}{\textit{e}.\textit{g}.}

\begin{document}

\title{Exploring Inconsistent Knowledge Distillation for Object Detection with Data Augmentation}

\author{Jiawei Liang}
\orcid{0000-0003-1143-6873}
\affiliation{%
  \institution{School of Computer Science and Engineering, Sun Yat-Sen University, China}
  \country{}
}
\email{liangjw57@mail2.sysu.edu.cn}

\author{Siyuan Liang}
\orcid{0000-0002-6154-0233}
\affiliation{
  \institution{State Key Laboratory of Information Security, Institute of Information Engineering, Chinese Academy of Sciences, China}
  \country{}
}
\email{liangsiyuan@iie.ac.cn}
\authornote{The corresponding author.}

\author{Aishan Liu}
\orcid{0000-0002-4224-1318}
\affiliation{
  \institution{\emph{NLSDE}, Beihang University, China}
  \institution{Institute of Dataspace, Hefei, China}
  \country{}
}
\email{liuaishan@buaa.edu.cn}
\authornotemark[1]

\author{Ke Ma}
\orcid{0000-0003-4178-0907}
\affiliation{
  \institution{School of Electronic, Electrical and Communication Engineering, University of  Chinese Academy of Sciences, China}
  \country{}
}
\email{make@ucas.ac.cn}

\author{Jingzhi Li}
\orcid{0000-0001-7054-9267}
\affiliation{
  \institution{State Key Laboratory of Information Security, Institute of Information Engineering, Chinese Academy of Sciences, China}
  \country{}
}
\email{lijingzhi@iie.ac.cn}

\author{Xiaochun Cao}
\orcid{0000-0001-7141-708X}
\affiliation{
  \institution{School of Cyber Science and Technology, Sun Yat-Sen University, China}
  \country{}
}
\email{caoxiaochun@mail.sysu.edu.cn}



\renewcommand{\shortauthors}{}
\newcommand{\method}{}

\begin{abstract}
Knowledge Distillation (KD) for object detection aims to train a compact detector by transferring knowledge from a teacher model. Since the teacher model perceives data in a way different from humans, existing KD methods only distill knowledge that is consistent with labels annotated by human expert while neglecting knowledge that is not consistent with human perception, which results in insufficient distillation and sub-optimal performance. In this paper, we propose inconsistent knowledge distillation (IKD), which aims to distill knowledge inherent in the teacher model's counter-intuitive perceptions. We start by considering the teacher model's counter-intuitive perceptions of frequency and non-robust features. Unlike previous works that exploit fine-grained features or introduce additional regularizations, we extract inconsistent knowledge by providing diverse input using data augmentation. Specifically, we propose a sample-specific data augmentation to transfer the teacher model's ability in capturing distinct frequency components and suggest an adversarial feature augmentation to extract the teacher model's perceptions of non-robust features in the data. Extensive experiments demonstrate the effectiveness of our method which outperforms state-of-the-art KD baselines on one-stage, two-stage and anchor-free object detectors (at most +1.0 mAP). Our codes will be made available at \url{https://github.com/JWLiang007/IKD.git}.
\end{abstract}




\maketitle

\begin{figure}[ht]
    \centering
    \includegraphics[width=0.45\textwidth]{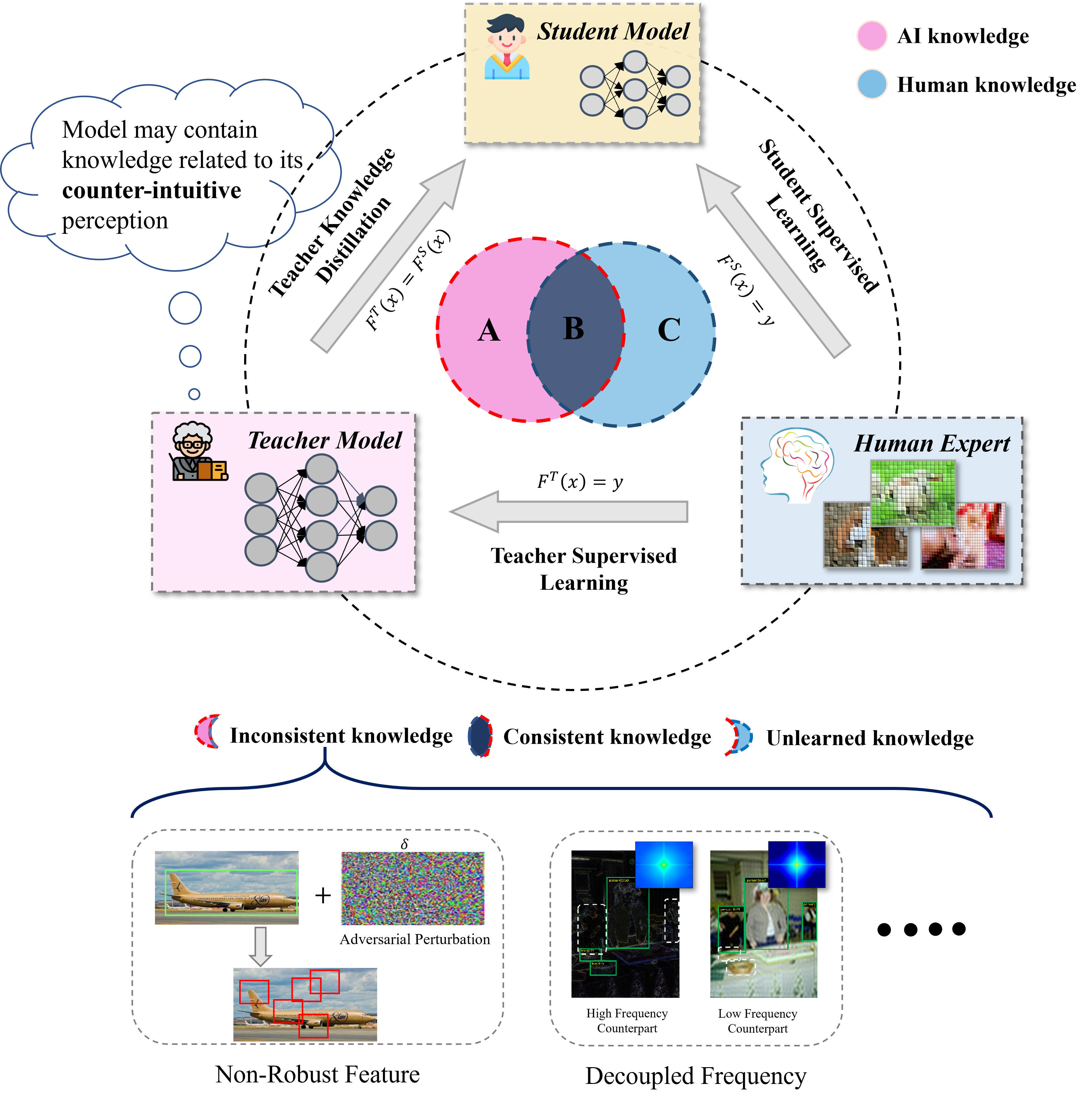}
	\caption{KD for object detection can be viewed as a combined distillation of AI knowledge and human knowledge. Since the AI model perceives the data differently from human, we hypothesize that the model may hold informative knowledge derived from its counter-intuitive perceptions and investigate inconsistent knowledge distillation.}
	\label{fig:main}
\end{figure}

\section{Introduction}

Deep-learning-based object detection has achieved remarkable progress~\cite{jiang2017face,zantalis2019review}, however, their parameters have also increased exponentially being eager for compression techniques~\cite{choudhary2020comprehensive} to help deployment on resources-limited devices. To solve this issue, knowledge distillation (KD)~\cite{hinton2015distilling} has been proposed to train a compact student model from a large teacher model with a similar structure and the same task goal.

As shown in Figure \ref{fig:main}, KD for object detection involves distillation of both AI knowledge from a teacher model and human knowledge from labels annotated by human expert. Current methods perform supervised learning to transfer human knowledge (illustrated as \textcolor{blue!85}{B}+\textcolor{cyan!65}{C}). For AI knowledge, current KD methods~\cite{wang2019distilling,zhang2020improve,yang2022focal} only focus on transferring consistent knowledge (denoted as \textcolor{blue!85}{B}) that aligns with labels annotated by human expert, \eg, these methods introduce additional regularizations to enforce distillation of more intuitive representations. However, recent studies~\cite{wang2020high,ilyas2019adversarial,yin2019fourier,bai2022improving,geirhos2020shortcut} suggest that convolutional neural networks (CNNs) can interpret image data differently from human perception, \eg, the model can exploit some patterns in the data that are not visible or comprehensible to humans to make predictions. These studies also attribute generalization behaviour of CNNs to such counter-intuitive perception. Motivated by this understanding, we argue that existing methods only transferring consistent knowledge result in insufficient distillation and propose to transfer knowledge relevant to the teahcer model's counter-intuitive perception, which we call inconsistent knowledge distillation  (denoted as \textcolor{red!95}{A}). By incorporating inconsistent knowledge distillation, we aim to transfer a broader range of knowledge from the teacher model, potentially improving the performance of the student model.

\begin{figure}
  \centering
  \subfigure[High frequency counterpart] {
    \label{fig: Detection Results of High frequency counterpart}
    \includegraphics[width=0.35\linewidth]{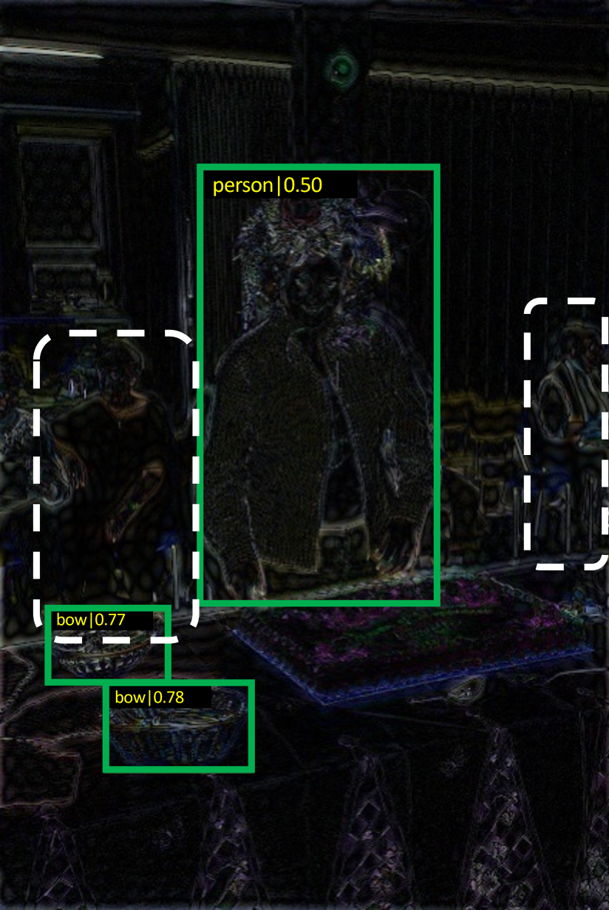}
  }\hspace{+5mm}
  \subfigure[Low frequency counterpart] {
    \label{fig: Detection Results of Low frequency counterpart}
    \includegraphics[width=0.35\linewidth]{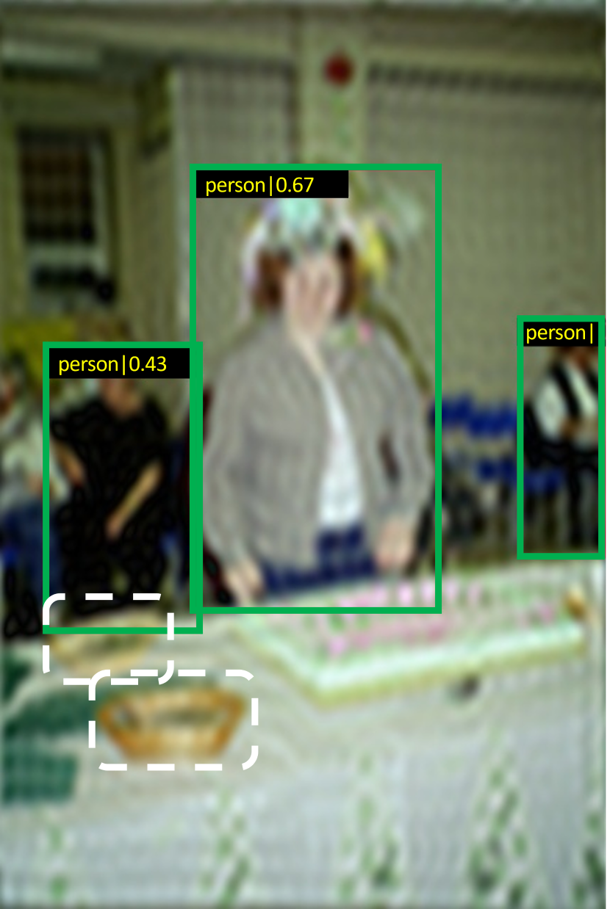}
  }
  \caption{Detection results of high and low frequency counterparts. White dashed bounding boxes indicate missed predictions. The object detector fails to detect small objects in the low frequency counterpart~\subref{fig: Detection Results of Low frequency counterpart} whereas it fails to detect larger objects in the high frequency counterpart~\subref{fig: Detection Results of High frequency counterpart}.}
  \label{fig: Detection Results of diff frequency counterparts}
\end{figure}

We mainly focus on two types of inconsistent knowledge for distillation. The first one is related to CNNs' ability in capturing distinct frequency components of image data. Prior work~\cite{wang2020high} suggests that, while humans mainly perceive low-frequency semantic components of an image, CNN-based object detectors can perceive both low and high frequency components. Additionally, as illustration in Figure~\ref{fig: Detection Results of diff frequency counterparts}, object detectors rely more on low frequency for large objects and rely more on high frequency components for small objects. This observation demonstrates CNNs' model-specific perceptions of distinct frequency components. The second one is related to CNNs' ability in capturing non-robust features of image data. Previous study~\cite{ilyas2019adversarial}
attributes the phenomenon of adversarial examples~\cite{wei2018transferable, liang2020efficient, liang2022parallel, liang2022large, liang2022imitated, liu2023x, wang2021dual, liu2019perceptual, liu2020bias, liu2020spatiotemporal, liu2022harnessing} to CNNs' perceptions of non-robust features, which are highly predictive but incomprehensible patterns derived from data. Despite adversarial vulnerability, non-robust features also reveal how CNNs perceive data in a counter-intuitive way.

Since the aforementioned inconsistent knowledge is relevant to multiple data patterns, we propose to distill inconsistent knowledge by providing diverse input using data augmentation. We propose a unified data augmentation framework. \textbf{Firstly}, since the teacher model relies on distinct frequency components when predicting on objects of different sizes, we propose a sample-specific data augmentation. Specifically, we enhance different frequency components of an image based on the size of objects within the image using different frequency augmentation. By frequency augmentation, we can encourage the teacher model to transfer inconsistent knowledge on distinct frequency components. \textbf{Secondly}, since adversarial examples can explicitly extract the teacher model's knowledge concerning non-robust features, we propose an adversarial feature augmentation that leverages adversarial examples alongside clean images as input and imitate the teacher model's deep adversarial features to mine inconsistent knowledge concerning non-robust features from the teacher detector.

 Our \textbf{contributions} can be summarized as follows.

\begin{itemize}
 
    \item We for the first time investigate distillation of inconsistent knowledge for object detection from data perspective and propose a general data augmentation framework.

    \item We propose a sample-specific data augmentation to distill the teacher model's inconsistent knowledge concerning frequency and an adversarial feature augmentation method to distill the teacher model's inconsistent knowledge concerning non-robust features.

    \item Extensive experiments show that our proposed general knowledge distillation framework could be easily extended to existing methods, and could outperform state-of-the-art KD methods for one-stage, two-stage and anchor-free object detectors by at most 1.0 mAP.

\end{itemize}

\begin{figure*}[ht]
    \centering
    \includegraphics[width=0.7\textwidth]{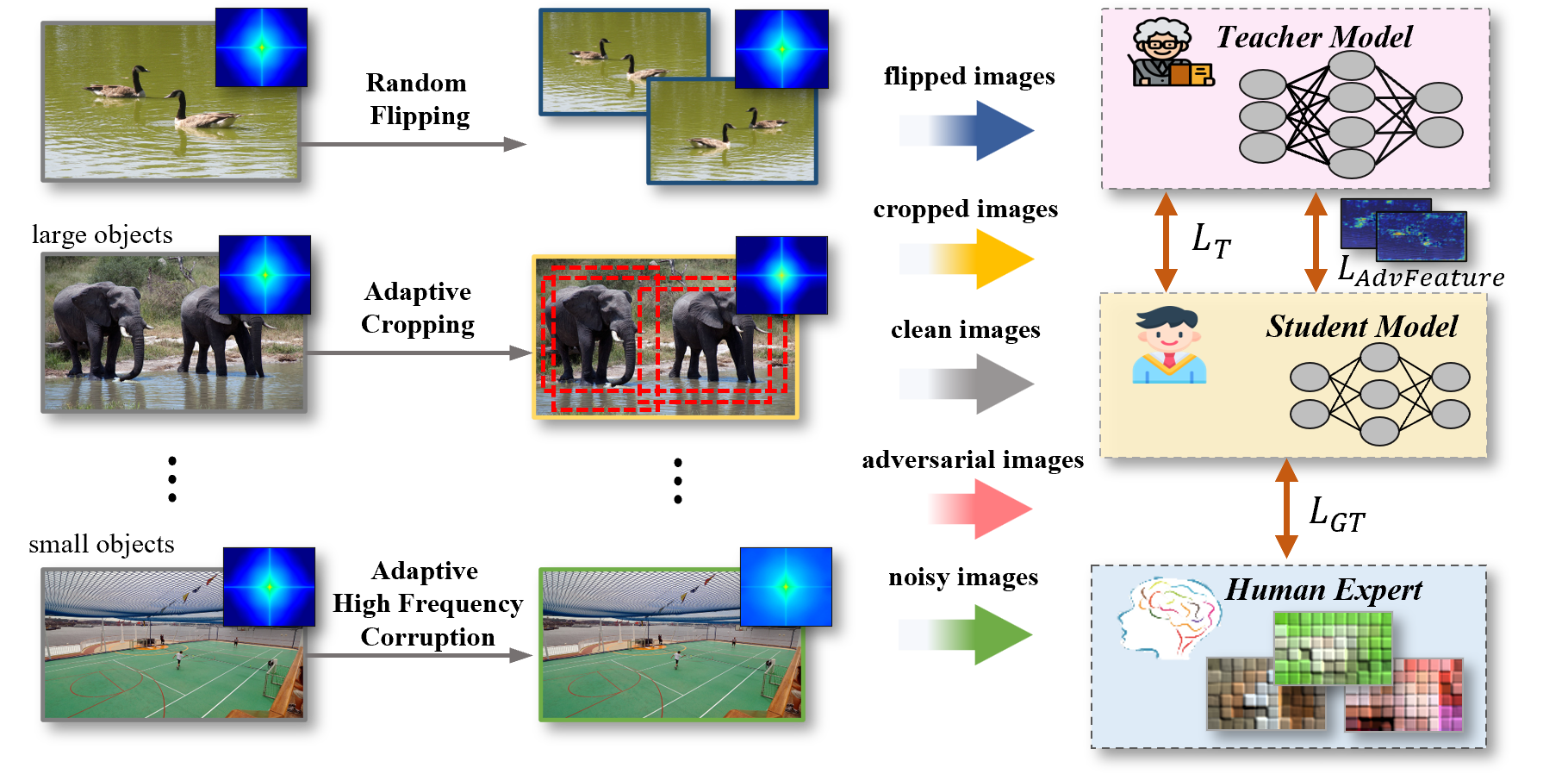}
	\caption{ Overall framework of our augmentation method. We propose a sample-specific data augmentation to distill inconsistent knowledge concerning frequency by enhancing distinct frequency components based on the size of objects. We propose an adversarial feature augmentation to distill inconsistent knowledge concerning non-robust features by imitating adversarial feature maps of the teacher model.}
	\label{fig:framework}
\end{figure*}

\section{Related work}

\subsection{Knowledge Distillation for Object Detection} 
Knowledge distillation is a commonly used method for object detection that compresses a large detector to a compact detector with a similar structure. Mimic network~\cite{chen2017learning} firstly proposed the knowledge distillation framework for object detection which optimizes the similarity in the same region on features coming from student and teacher networks. To better use the spatial information of objects, \cite{wang2019distilling} utilized the local near object regions to estimate imitation regions and imitated fine-grained features. Compared to spatial distillation methods, many methods integrated the channel information into the distillation framework. \cite{zhang2020improve} used both channel and spatial attention masks to encourage students to mimic regions with higher activation and introduced non-local modules to capture global information. \cite{shu2021channel} aligned different weights of each channel to transfer more channel-wise knowledge. \cite{yang2022focal} pointed out that equally distilling different regions may harm the performance, thus they separated the foreground and the background with different weights using spatial and channel attention masks. MGD ~\cite{yang2022masked} abnegated spatial and channel priors and trained an adversarial generator, learning the consistency between the erased features of the student and the features of the teacher. 

In summary, the majority of existing works use additional regularizations to encourage the model to distill fine-grained features that align with ground truth. However, they tend to overlook the fact that the model also contains inconsistent knowledge.

\subsection{Data Augmentation} 

Data augmentation mitigates data starvation or overfitting of network learning by selecting or constructing similar data from source training data. We can classify existing data augmentation methods into four broad categories~\cite{kaur2021data} as follows: (1) geometric transformations such as rotation, flipping, and translation, which are commonly used in the deep neural network~\cite{krizhevsky2012imagenet}; (2) photometric transformations that modify the pixel or intensity values~\cite{simonyan2014very} ; (3)  random occlusion techniques such as random erasing~\cite{zhong2020random}, gird mask~\cite{chen2020gridmask} and cutout~\cite{devries2017improved}, which enable the model to resist object occlusions; and (4) deep learning based techniques such as adversarial training~\cite{bagheri2018adversarial} and generative adversarial networks~\cite{mansourifar2019virtual,tanaka2019data}. 

Existing KD methods for object detection overlook the importance of data augmentation and simply use the same data augmentation methods used in object detection task, \eg, padding, flipping. In this paper, we demonstrate that data augmentation can be a powerful tool for inconsistent knowledge distillation.

\section{Methodology}
Firstly, we formulate KD for object detection and demonstrate the deficiencies of current KD methods in distilling inconsistent knowledge. Secondly, we introduce our data augmentation framework used to distill inconsistent knowledge, which involves a sample-specific data augmentation method and an adversarial feature augmentation method. Finally, we show the algorithm flow.

\subsection{Formulation of Knowledge Distillation for Object Detection} 
The goal of knowledge distillation (KD) for object detection is to train a student object detector by mimicking the features or logits of a larger teacher object detector together with supervision of the ground truth. The general loss function for knowledge distillation can be represented as follow:
\begin{equation}
	\label{KD loss}
    L_{KD}(\mathbf{x}) = L_{GT}(\mathbf{x}) + \alpha \cdot L_{T}(\mathbf{x}),
\end{equation}
where $L_{GT}$ represents the task loss of object detection and $L_{T}$ refers to the distillation loss that forces the student model to imitate deep representations or logits of the teacher model. $\alpha$ is a hyper-parameter that balances the two losses. In this paper, we focus on the feature-based KD methods for object detection. Then the distillation loss $L_{T}$ is defined as:
\begin{equation}
	\label{distillation loss}
    L_{T}(\mathbf{x}) = L_{feat}(F^{S}(\mathbf{x}) - F^{T}(\mathbf{x})) ,
\end{equation}
where $F^{S}(\mathbf{x})$ and $F^{T}(\mathbf{x})$ refer to deep features of the student and teacher models respectively. $L_{feat}$ represents the loss function that minimizes the discrepancy between features of the student and teacher models.

Existing KD methods for object detection utilize the same data augmentation as the teacher model, \eg, flipping, which we denote as $a_T(\cdot)$. We argue that using the same data augmentation $a_T(\cdot)$ in KD can not adequately distill knowledge from the teacher model. Firstly, the teacher model is trained with:
\begin{equation}
	\label{task loss for teacher}
    \theta^{T} = \mathop{\arg\min}\limits_{\theta} L^{T}_{task}(H^{T}(a_{T}(\mathbf{x}),\theta), y)
\end{equation}
where $H^{T}(a_{T}(\mathbf{x}),\theta)$ refers to the output of the teacher model with augmented $a_T(\mathbf{x})$ as input, $L^{T}_{task}$ represents the task loss of the teacher model, and $y$ is the ground truth. The equation~\ref{task loss for teacher} reveals that the teacher model is trained to fit well with the ground truth under the data augmentation $a_{T}(\cdot)$. In other word, if we use the same $a_{T}(\cdot)$ in KD, it will encourage distillation of the teacher model's knowledge that coincides with the ground truth, \ie, consistent knowledge (\textcolor{blue!85}{B} in figure~\ref{fig:main}). However, previous works~\cite{wang2020high,ilyas2019adversarial,yin2019fourier,bai2022improving} suggest that CNN-based models could make predictions in a counter-intuitive way despite being trained with supervision of human annotated labels. These studies inspire us that the teacher model may contain knowledge that explains its superior performance, \ie, inconsistent knowledge (\textcolor{red!95}{A} in figure~\ref{fig:main}). Therefore, in the following sections, we will delve into two types of available inconsistent knowledge and introduce our methods to distill inconsistent knowledge.

\subsection{Sample-Specific Data Augmentation from Fourier Perspective} \label{Section: Sample-Specific Data Augmentation}
Earlier work~\cite{wang2020high} formulates the perceptional disparity between CNN-based models and human from a frequency perspective. It posits that CNN-based models can exploit high frequency components of input images beyond low frequency components which human annotated labels encourage the model to focus on. One intuitive interpretation is illustrated in Figure~\ref{fig: Detection Results of diff frequency counterparts}. We use a CNN-based object detector RetinaNet~\cite{lin2017focal} to predict on low and high frequency counterparts of an image. The results show that, although low frequency counterpart of the image (in Figure~\ref{fig: Detection Results of Low frequency counterpart}) is more discriminative for human as compared to high frequency counterpart (in Figure~\ref{fig: Detection Results of High frequency counterpart}), the model does not always perform better on low frequency counterpart, particularly for small objects. The results suggest : (1) the model utilize both low and high frequency components for predictions; (2) the model would utilize distinct frequency components based on the scale of instances within an image.

The above observations inspire us to propose a sample-specific data augmentation method based on the scale of instances within an image in order to distill the teacher model's perceptions of distinct frequency components. Concretely, for images where small objects are dominant, the model would focus more on high frequency components. Thus we would enhance the high frequency components for this image to distill the teacher model's perceptions of high frequency components. On the contrary, for images where large objects are dominant, we would enhance the low frequency components. Now that we have our basic data augmentation strategy, we need to specify what data augmentation to use. In the following, we provide an analysis of different data augmentation methods from Fourier perspective.  

\begin{figure}
    \centering
      \subfigure[Original] {
    \label{fig: Fourier Spectrum of Clean Images}
    \includegraphics[width=0.40\linewidth]{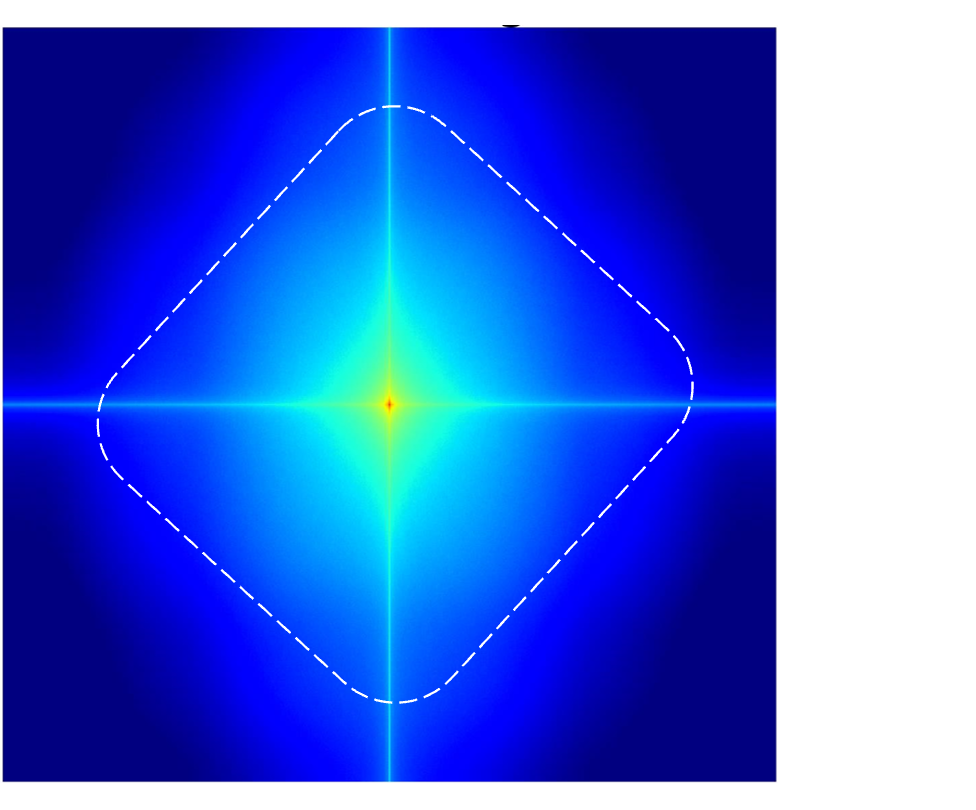}
    }
      \subfigure[Flipping] {
    \label{fig: Fourier Spectrum of Flipped Images}
    \includegraphics[width=0.40\linewidth]{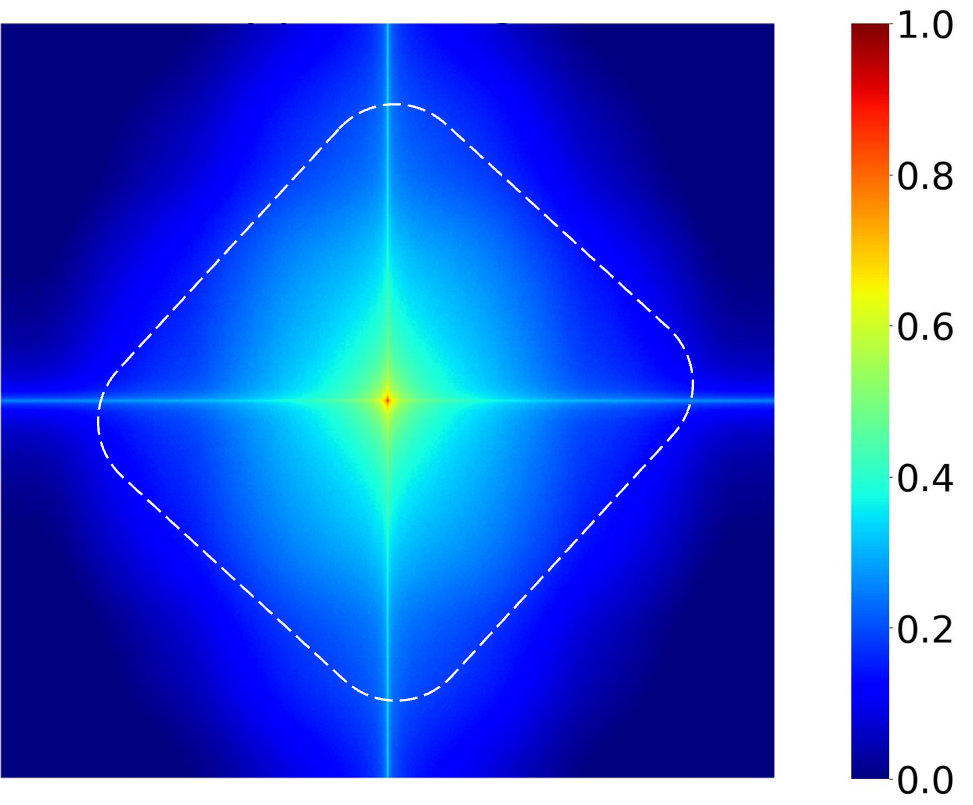}}
      \subfigure[Cropping] {
    \label{fig: Fourier Spectrum of Cropped Images}
    \includegraphics[width=0.40\linewidth]{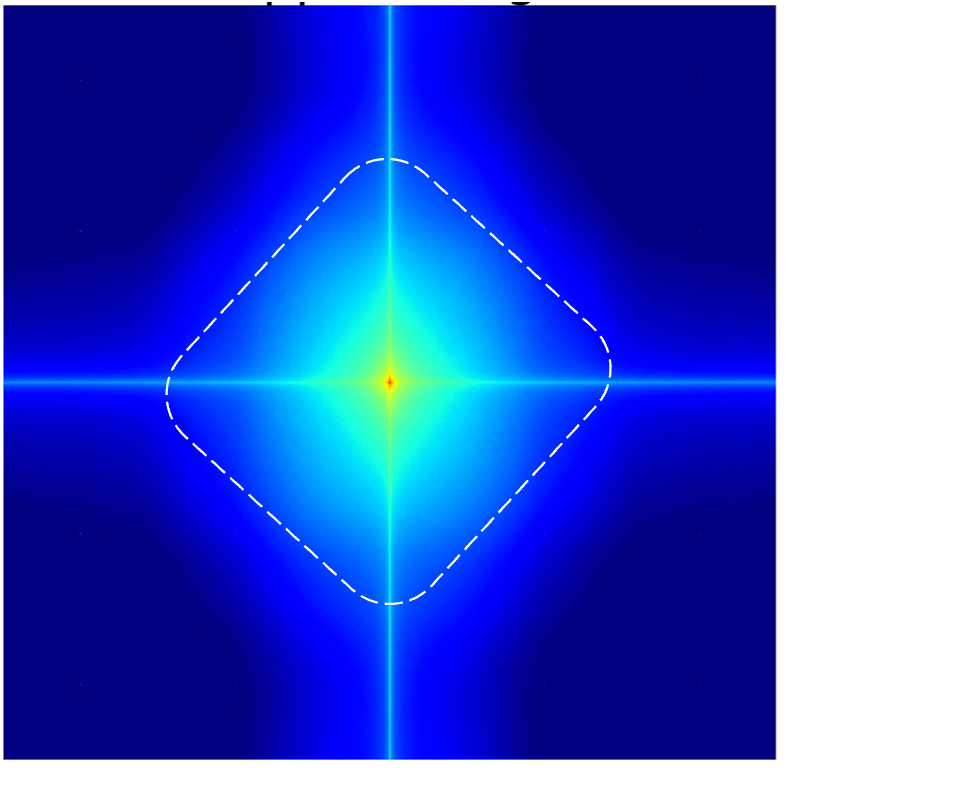}
    }
      \subfigure[Gaussian Noise Corruption] {
    \label{fig: Fourier Spectrum of Gaussian noise corrupted Images}
    \includegraphics[width=0.40\linewidth]{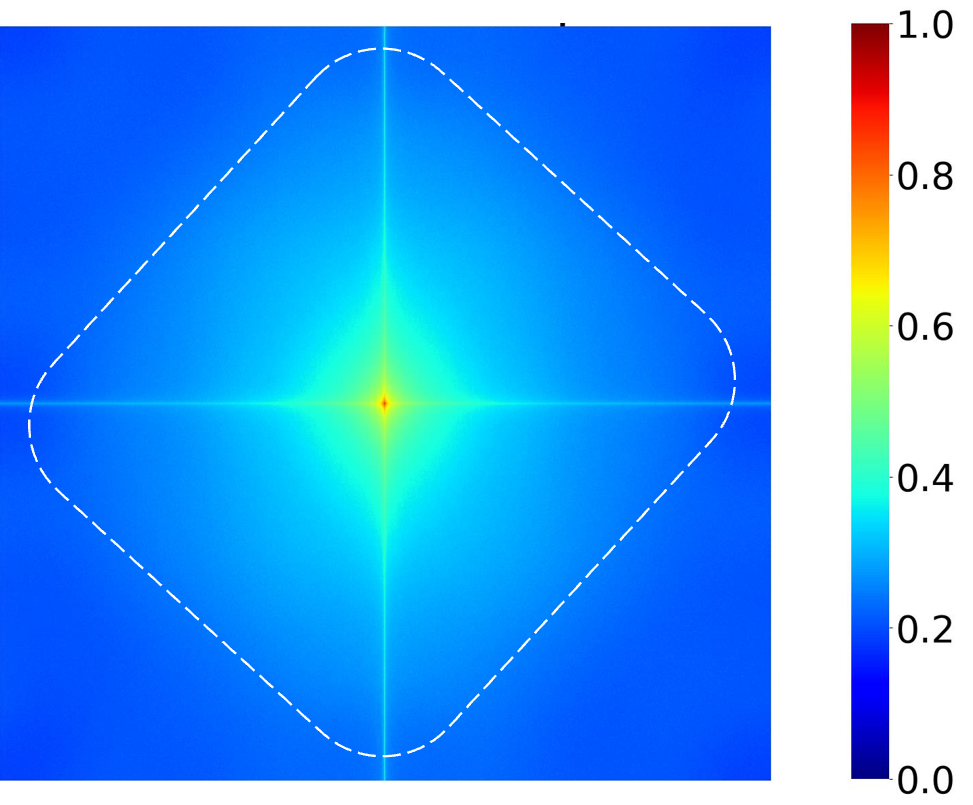}}
	\caption{Fourier spectrum of clean images~\subref{fig: Fourier Spectrum of Clean Images} and images augmented with flipping~\subref{fig: Fourier Spectrum of Flipped Images}, cropping~\subref{fig: Fourier Spectrum of Cropped Images} and Gaussian noise corruption~\subref{fig: Fourier Spectrum of Gaussian noise corrupted Images}, respectively. Flipping does not modify frequency components. Cropping makes images more concentrated in low frequency whereas Gaussian noise introduces more high frequency components compared to clean images.}

	\label{fig:fourier_spectrum}
\end{figure}

\textbf{Data Augmentation from Fourier Perspective:}
Inspired by \cite{yin2019fourier}, we analyze three representative augmentation methods from Fourier perspective \ie, random flipping, random cropping, and Gaussian noise corruption. To obtain the frequency characteristics of the three augmentation methods, we apply the 2D discrete Fourier transform (DFT) to images augmented by the three methods respectively and clean images for comparison. Specifically, we obtain each Fourier spectrum by averaging over a collection of images with a batch size of 256. Then, we get the Fourier spectrum of each type of image and visualize them as heatmaps. 

In Figure \ref{fig:fourier_spectrum}, we can draw several observations: (1) random flipped images exhibit a similar Fourier spectrum to clean images. (2) Compared to clean images, randomly cropped images have a higher concentration at the center of the Fourier spectrum, which corresponds to low frequency components; (3) On the contrary, images with Gaussian noise corruption have higher values at points away from the center which indicates high-frequency components. Overall, we find that different augmentation methods bring distinct changes to the frequency statistics of an image.

Here, together with our observations from Fourier perspective, we formulate our sample-specific data augmentation method strategy regarding object scale as follows: 
\begin{equation}
a_{data}(\mathbf{x})=\left\{
	\begin{aligned}
	&a_1(\textbf{x}), \quad \exists \mathbf{b} \in\{\rm{GT}_{\mathbf{x}}\}, p(\mathbf{b}) \geq \tau\\
	&a_2(\textbf{x}), \quad \forall \mathbf{b} \in\{\rm{GT}_{\mathbf{x}}\}, p(\mathbf{b}) < \tau \\
	\end{aligned}
	\right
	.
    \label{equ: data augmentation}
\end{equation}
where p$(\mathbf{b})$ is the area of the bounding box from ground-truth (GT) of the image $\mathbf{x}$. $a_1$ and $a_2$ respectively represent the \textbf{adaptive cropping} and \textbf{adaptive high frequency corruption}. $\tau$ represents the area threshold that defines the small object.

For images without small objects, we apply cropping to images to enhance low frequency components. Since the model focus more on low frequency components for large objects, by using cropping, we can efficiently extract the teacher model's perceptions of low-frequency components and improve the performance on large objects. Specifically, we randomly crop the edges of images with a certain relative ratio. 

For images dominated by small objects, we adds high frequency noises to clean images to enhance high frequency components. Since the model focus more on high frequency components for small objects, this method efficiently extracts the teacher model's perceptions of high frequency components and improve the performance on small objects. Specifically, we use Gaussian noise sampled from $N(0,\sigma^2)$ as the high frequency corruption and add the noise to clean images with a certain probability. 

\subsection{Adversarial Feature Augmentation} 
Most previous feature-based KD methods focus on distilling consistent knowledge from the teacher model out of consideration of consistency between the output of the teacher model and the ground truth. It can be expressed as $H^T(\mathbf{x}) = y$ where $H^T(\mathbf{x})$ denotes the prediction of teacher model and $y$ is ground truth. 

However, the distillation loss in Equation~\ref{KD loss} only enforces the student model to imitate deep representations of the teacher model, such that $F^T_i(\mathbf{x}) = F^S_i(\mathbf{x})$, where $i=1,2,...,K$ and $F^T_i(\mathbf{x})$, $F^S_i(\mathbf{x})$ denote the $i^{th}$ deep representations of the teacher and student models with $\mathbf{x}$ as input. This implies that the distillation of knowledge from the teacher model may be limited by the constraint of $H^T(\mathbf{x}) = y$. Therefore, we hypothesize that the feature-based knowledge distillation is less restrictive on input and allows $H^T(\mathbf{x}) \neq y$. Moreover, previous study~\cite{ilyas2019adversarial} suggests that CNN-based models would utilize non-robust features for predictions, which are highly predictive but imperceptible patterns induced from input data. Although the model's sensitivity to non-robust features leads to adversarial vulnerability~\cite{ma2022tale, ma2021poisoning, jia2022adversarial, jia2022prior, zhang2021interpreting, tang2021robustart, liu2021training, liu2023exploring, guo2023towards}, such features also reveals how the model perceives data differently from human. This inspire us to distill the model's perceptions of non-robust features. Since the adversarial examples depend on non-robust features, we can use adversarial examples to extract the model's knowledge concerning non-robust features.

Therefore, we propose an adversarial feature augmentation method to mine more inconsistent knowledge from the teacher model. Concretely, we take adversarial examples alongside with clean images as input and enforce the student model to imitate the deep adversarial representations of the teacher model. We generate adversarial examples using a transfer-based attack method and take the teacher model as the victim model. We formulate adversarial examples $ a_{adv}(\mathbf{x}) = \mathbf{x} + \mathbf{\delta}_{t}$ as follows:
\begin{equation}
\label{adversaril attack}
   H^{T}(a_{adv}(\mathbf{x})) \neq H^{T}(\mathbf{x}) \quad s.t., ||a_{adv}(\mathbf{x})-\mathbf{x}||_{\infty} \leq \epsilon, 
\end{equation}
where $t$ is the number of iteration, and $\delta$ denotes the adversarial perturbation. $H^{T}$ represents the teacher model. $\epsilon$ represents the maximal magnitude of the adversarial perturbation.

Then we calculate the adversarial feature loss as follows:
\begin{equation}
L_{AdvFeature}(\hat{\mathbf{x}})=\sum_{i=1}^K || F^T_{i}(\hat{\mathbf{x}}) - F^S_{i}(\hat{\mathbf{x}}) ||_2,
\label{equ: adversarial feature loss}
\end{equation}
where $\hat{\mathbf{x}}=a_{adv}(\mathbf{x})$ is adversarial example defined by Equation~\ref{adversaril attack}. $F^T_{i}(\hat{\mathbf{x}})$, $F^S_{i}(\hat{\mathbf{x}})$ denote the $i^{th}$ deep feature maps of the teacher and student models with an adversarial example as input. $K$ represents the total number of the feature maps. Specifically, for pipeline with adversarial examples as input, we only minimize the distillation loss and do not use supervision of the ground truth. This is because our goal is to imitate deep representations of the teacher model rather than correct the predictions.

\subsection{Unified Data and feature augmentation}
To sum up, we propose a sample-specific data augmentation method based on objects scales and a feature augmentation method. The overall knowledge distillation loss can be formulated as:
\begin{equation}
L_{KD} = L_{GT} + \alpha \cdot L_{T} + \beta \cdot L_{AdvFeature}.
\label{equ: overall loss}
\end{equation}

The overall procedure of our method is detailed in Algorithm \ref{procedure}.

\begin{algorithm}[t]
    \caption{Data and Feature Augmentation}
    \begin{algorithmic}[1]
        \REQUIRE Teacher Detector: $T$, Student Detector: $S$, Input: $\mathbf{x}$, label: $y$, hyper-parameter: $\beta$ \\
        \ENSURE $S$ \\ 
        \STATE Using DIFGSM~\cite{xie2019improving} to generate adversarial examples $a_{adv}(\mathbf{x})$ in Equation~\ref{adversaril attack} of $\mathbf{x}$ with $T$ as victim model
        \STATE Applying data augmentation in Equation~\ref{equ: data augmentation} to $\mathbf{x}$ regarding object scale to get $a_{data}(\mathbf{x})$
        \STATE Using $S$ to get clean features $\{F^{S}_{1},...F^{S}_{i},...\}^{clean}$ and output $\hat{y}^{S}_{clean}$ of input $a_{data}(\mathbf{x})$ and get adversarial features $\{F^{S}_{1},...F^{S}_{i},...\}^{adv}$ of input $a_{adv}(\mathbf{x})$
        \STATE Using $T$ to get clean features $\{F^{T}_{1},...F^{T}_{i},...\}^{clean}$ of input  $a_{data}(\mathbf{x})$ and get adversarial features $\{F^{T}_{1},...F^{T}_{i},...\}^{adv}$ of input $a_{adv}(\mathbf{x})$
        \STATE Calculating the task loss of the model: $L_{GT}$
        \STATE Calculating the clean features distillation loss by arbitrary object detection distillation algorithm: $L_{T}$
        \STATE Calculating the adversarial feature distillation loss $L_{AdvFeature}$ in Equation~\ref{equ: adversarial feature loss}
        \STATE Using the overall loss $L_{KD}$ in Equation~\ref{equ: overall loss} to update $S$ 
    \end{algorithmic}
    \label{procedure}
\end{algorithm}

\begin{table*}[t]
    \centering
    \begin{tabular}{c|c|c|llll}
        \toprule
        Teacher&Student &Method & mAP &  mAP$_S$ &  mAP$_M$ &  mAP$_L$  \\
        \midrule
        
        \multirow{7}{*}{\makecell[c]{RetinaNet \\ ResNeXt101 \\(41.0) } } &
        \multirow{7}{*}{\makecell[c]{RetinaNet \\Res50 (37.4) }} 
        & Student   & 37.4   & 20.6    & 40.7 & 49.7    \\
        
        & & FKD~\cite{zhang2020improve} &39.6 (+2.2) & 22.7 & 43.3 & 52.5 \\
        & & CWD~\cite{shu2021channel}  &40.8 (+3.4) &22.7 &44.5 &55.3 \\
        & & FGD~\cite{yang2022focal} & 40.7 (+3.3) & 22.9& 45.0& 54.7 \\
        & & MGD~\cite{yang2022masked} &41.0 (+3.6)&23.3 &45.0 &55.4 \\
        & & Ours (+MGD)& \textbf{41.1 (+3.7)} & 23.6 &	45.4 &	54.8 \\
        & & Ours (+FGD)& \textbf{41.2 (+3.8)} & 23.9 & 45.4 & 54.9\\

        \midrule
        \multirow{7}{*}{\makecell[c]{Cascade \\ Mask RCNN \\ ResNeXt101 \\(47.3) } } &
        \multirow{7}{*}{\makecell[c]{Mask RCNN \\ Res50 (39.2) }} 
        & Student   & 39.2   & 22.9 & 42.6 & 51.2     \\
        
        & & FKD~\cite{zhang2020improve} & 41.7 (+2.5) & 23.4 & 45.3 & 55.8 \\
        & & CWD~\cite{shu2021channel}  &41.2 (+2.0) & 23.1 & 45.4 & 54.7  \\ 
        & & FGD~\cite{yang2022focal} & 42.1 (+2.9) & 23.7& 46.2& 55.7\\
        & & MGD~\cite{yang2022masked} &42.3 (+3.1)&23.6 &46.3 &56.4 \\
        & & Ours (+FGD)& \textbf{42.4 (+3.2)} &24.7&46.7&55.9 \\
        & & Ours (+MGD)& \textbf{42.5 (+3.3)} &23.6 &	46.8 &	56.4\\
        \midrule
        \multirow{7}{*}{\makecell{RepPoints\\ResNeXt101\\(44.2)}} &
        \multirow{7}{*}{\makecell[c]{RepPoints\\ Res50 (38.6) }} 
        &Student & 38.6&22.5&42.2&50.4\\
        
        &&FKD~\cite{zhang2020improve} & 40.6 (+2.0)&23.4&44.6&53.0\\
        &&CWD~\cite{shu2021channel}&42.0 (+3.4)&24.1&46.1&55.0\\
        &&FGD~\cite{yang2022focal} &42.0 (+3.4)&24.0&45.7&55.6\\
        &&MGD~\cite{yang2022masked} & 42.3 (+3.7)&24.4&46.2&55.9\\
        & & Ours (+FGD)& \textbf{43.0 (+4.4)} & 25.7 & 47.4 & 56.4 \\ 
        & & Ours (+MGD)& \textbf{43.0 (+4.4)} & 25.2 & 47.3 & 56.9 \\
        
        \bottomrule
    \end{tabular}
    \caption{Results of different KD methods evaluated on COCO dataset. We conduct experiments on one-stage, two-stage, and anchor-free object detectors. The student rows represent the results of training from scratch. We evaluate the effectiveness of our method in conjunction with FGD and MGD methods and present the results in Ours (+FGD) and Ours (+MGD) rows. }
    \label{comparison with KD baselines}
\end{table*}

\section{Experiments}

\subsection{Dataset and Evaluation}
We evaluate our method on the COCO2017 dataset~\cite{lin2014microsoft}, which contains 80 object categories. We use the 120k train images for training and 5k val images for testing. The performances of different models are evaluated in mean Average Precision (mAP) using MS COCO’s evaluation metrics~\cite{lin2014microsoft}. 

\subsection{Implementation Details}
We conduct our experiments on 8 A100 GPUs based on mmdetection~\cite{chen2019mmdetection} framework. To verify our approach, we apply our augmentation strategy to two previous feature-based KD methods, FGD~\cite{yang2022focal} and MGD~\cite{yang2022masked}. We adopt the original experiment setting and hyper-parameters for these two methods. We perform experiments on both one-stage, two-stage and anchor-free object detectors. We use ResNet-50~\cite{he2016deep} backbone for student object detectors and ResNeXt101~\cite{xie2017aggregated} backbone for teacher object detectors. For feature augmentation, we use DIFGSM~\cite{xie2019improving} to generate adversarial examples in an offline manner with maximum perturbation \(\epsilon = 8\) , number of iteration \(t = 5\), and step size \(\lambda = 2\). For data augmentation, we set the probability of adding high frequency corruption \(p = 0.3\), relative cropped ratio \(r = 0.2\). For feature augmentation, we set \(\beta = 1\times10^{-5}\) for one-stage and anchor-free object detectors and \(\beta = 2.5\times10^{-7} \) for two-stage object detectors. All models are trained for 24 epochs. Inspired by \cite{cho2019efficacy}, We use an early stopping strategy for our augmentation method, \ie, applying our augmentation method only in the first 16 epochs.

\begin{table*}
\small
    \centering
    \begin{tabular}{c|ccc|cccc}
        \toprule
        Index &  AdpHFC &  AdpCrop & AdvFeat  &  mAP  &  mAP$_S$ & mAP$_M$&  mAP$_L$  \\
        \midrule
        1   &            &           &            &40.3 	&	22.5 &44.3 &	54.2    \\
        2   &\checkmark  &           &            &40.3 &	23.6 &	44.5 &	54.6     \\
        3   &            &\checkmark &            &40.6 &	23.1 &	45.0 &	54.2      \\
        4   &\checkmark  &\checkmark &            &40.7 	&23.3 &	45.3 &	54.1    \\
        5   &            &   &   \checkmark      &40.6 &	22.7 &	44.7 &	54.2    \\
        6   &\checkmark  &\checkmark &\checkmark  & \textbf{40.9}& 23.4 &	45.4 & 54.3      \\

        \bottomrule
    \end{tabular}
    \caption{Ablation study to evaluate the effectiveness of the proposed data augmentation strategy and the feature augmentation method. We conduct our experiments with FGD~\protect\cite{yang2022focal} on RetinaNet~\cite{lin2017focal} object detector without extra data augmentation.}
    \label{ablation study}
\end{table*}

\subsection{Main Results}

We conduct experiments on three representative object detectors, \ie, one-stage object detector RetinaNet~\cite{lin2017focal}, two-stage object detector Mask RCNN~\cite{he2017mask} and anchor-free object detector RepPoints~\cite{yang2019reppoints}. To validate the effectiveness of our method, we apply our data augmentation method to two state-of-the-art KD methods for object detection, \ie, FGD~\protect\cite{yang2022focal} and MGD~\cite{yang2022masked}. We evaluate the performance of our method and make a thorough comparison with current KD methods for object detection, which is depicted in table~\ref{comparison with KD baselines}. \par
\textbf{Firstly}, for all three object detectors, our data augmentation method can further enhance the performance of current KD methods for object detection and achieve the state-of-the-art performance, \eg, our method yields a performance gain of 1.0 mAP for FGD on the RepPoints object detectors and renders a performance gain of 4.4 mAP compared to the student model trained from scratch, achieving the highest mAP of 43.0. These results demonstrate the generalizability and effectiveness of our proposed method. \textbf{Secondly}, our method yields less mAP gain for RetinaNet object detectors, which we attribute to the limit of the teacher model's performance. Since together with either FGD or MGD, our method surpasses the teacher model. \textbf{Thirdly}, we observe that our method provides less mAP gain for MGD compared to FGD. We suggest that this may be because MGD uses GAN~\cite{goodfellow2014generative} to reconstruct the student model's features, which can be viewed as another deep-neural-network-based augmentation. Thus together with GAN used in MGD, our augmentation method yields less mAP gain.

\begin{table}
\small
    \centering
        \begin{tabular}{c|c|cccc}
        \toprule        
        Index &  method &  mAP & mAP$_S$ &   mAP$_M$  & mAP$_L$ \\
        \midrule
        1   &       No Aug     &      40.3 &22.5 &44.3&54.2  \\
        2   &   cutout~\cite{devries2017improved}   & 40.6 &  22.8  & 44.9 &  54.4   \\
        3   &    copypaste~\cite{ghiasi2021simple}  & 40.6 &  23.8 & 45.2 & 53.9   \\
        4  & flipping & 40.7& 22.9& 45.0& 54.7  \\
        5  & Ours & 40.9& 23.4 & 45.4 & 54.3   \\
        6  & Ours (+flipping) & \textbf{41.2} & 23.9 & 45.4 & 54.9   \\
        \bottomrule
        \end{tabular}
    \caption{Comparisons with other data augmentation methods. No Aug means training without data augmentation.}
    \label{Comparisons with other data augmentation}
\end{table}

\subsection{Comparisons with Other Data Augmentation}
In this section, we present a comparison of our method with other popular data augmentation methods, \ie, flipping, cutout, and copypaste. We conduct experiments on the one-stage detector RetinaNet with FGD, and the results are depicted in Table \ref{Comparisons with other data augmentation}. Firstly, we observe that all the data augmentation methods improve the performance of knowledge distillation, indicating the significance of data augmentation in knowledge distillation. Secondly, our method outperforms the other data augmentation methods, achieving a 0.6 mAP gain over the result without data augmentation. This improvement validates the effectiveness of our method in improving knowledge distillation for object detection. Finally, we combine our method with the commonly used data augmentation method, \ie, flipping, and obtain the highest mAP. This result suggests that there is potential in further improvement of knowledge distillation by combining our method with other data augmentation methods.

\subsection{Ablation Studies}
In this section, we conduct ablation experiments to investigate the effectiveness of the proposed sample-specific data augmentation strategy and the feature augmentation strategy. Specifically, to evaluate our augmentation methods, we conduct ablation studies based on FGD without flipping. \par
We conduct several experiments as follows: (1) training with adaptive high frequency corruption, \ie, adpHFC;  (2) training with adaptive cropping, \ie, AdpCrop; (3) training with sample-specific data augmentation, \ie, adpHFC, AdpCrop; (4) training with adversarial feature augmentation, \ie, FeatAug; (5) training with both data and feature augmentation, \ie, adpHFC, AdpCrop, FeatAug. \par
As shown in Table \ref{ablation study}, we can draw some observations regarding performance improvement of the baseline when applying our augmentation method as follow: (1) training with adaptive high frequency corruption can significantly increase the mAP of small objects from 22.5 to 23.6 compared to that of medium-sized objects and large objects, which is strong evidence that enhancing high frequency components can help detectors perform better on small objects; (2) training with adaptive cropping mainly improves performance on medium-sized objects and brings 0.7 mAP gain for mAP$_M$ and 0.3 mAP gain for overall mAP from 40.3 to 40.6. Specifically, in this experiment, we set the threshold of the object size to be $32*32$, which means only input with medium-sized or large objects will be cropped. The improvement verifies our assumption; (3) training with feature augmentation can also boost the performance of knowledge distillation with mAP increase of 0.3, in a way that strengthens the feature distillation process; (4) combinations of our augmentation strategies can further improve the performance. It is worth noticing that we achieve the best result when combining all augmentation strategies and boost the overall mAP to 40.9 which outperforms the baseline by a large margin. \par
To sum up, our ablation studies demonstrate that both adaptive data augmentation can boost the performance and they function separately on objects with different scales; feature augmentation can also bring performance gain for knowledge distillation. 

\begin{figure*}
  \centering
  \subfigure[Teacher Model] {
    \label{fig: Detection Results of Teacher Model}
    \includegraphics[width=0.20\linewidth]{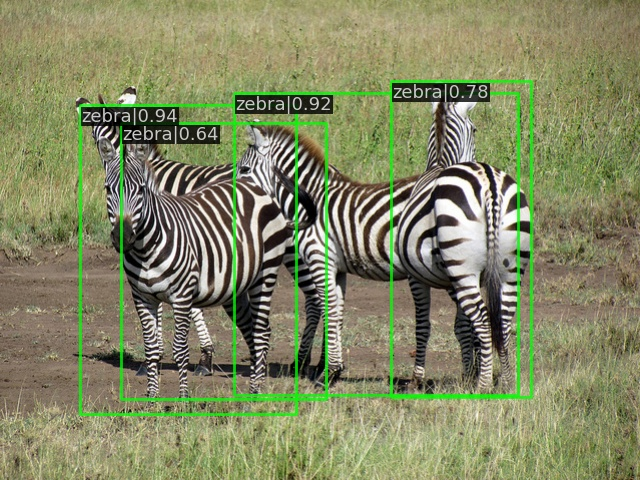}
  }
  \subfigure[Ours (+FGD)] {
    \label{fig: Detection Results of DFA}
    \includegraphics[width=0.20\linewidth]{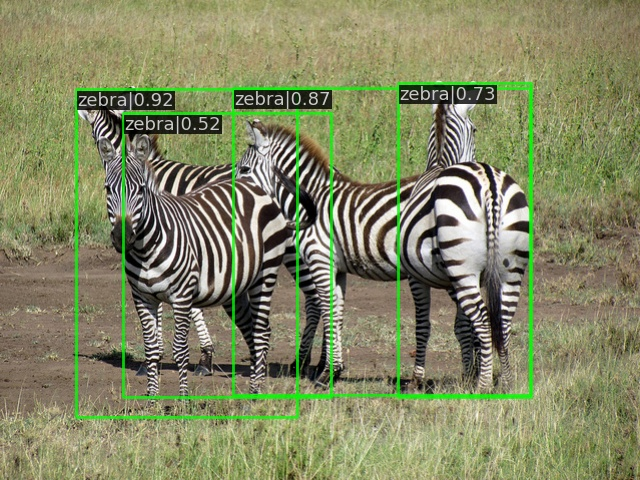}
  }
  \subfigure[FGD] {
    \label{fig: Detection Results of FGD}
    \includegraphics[width=0.20\linewidth]{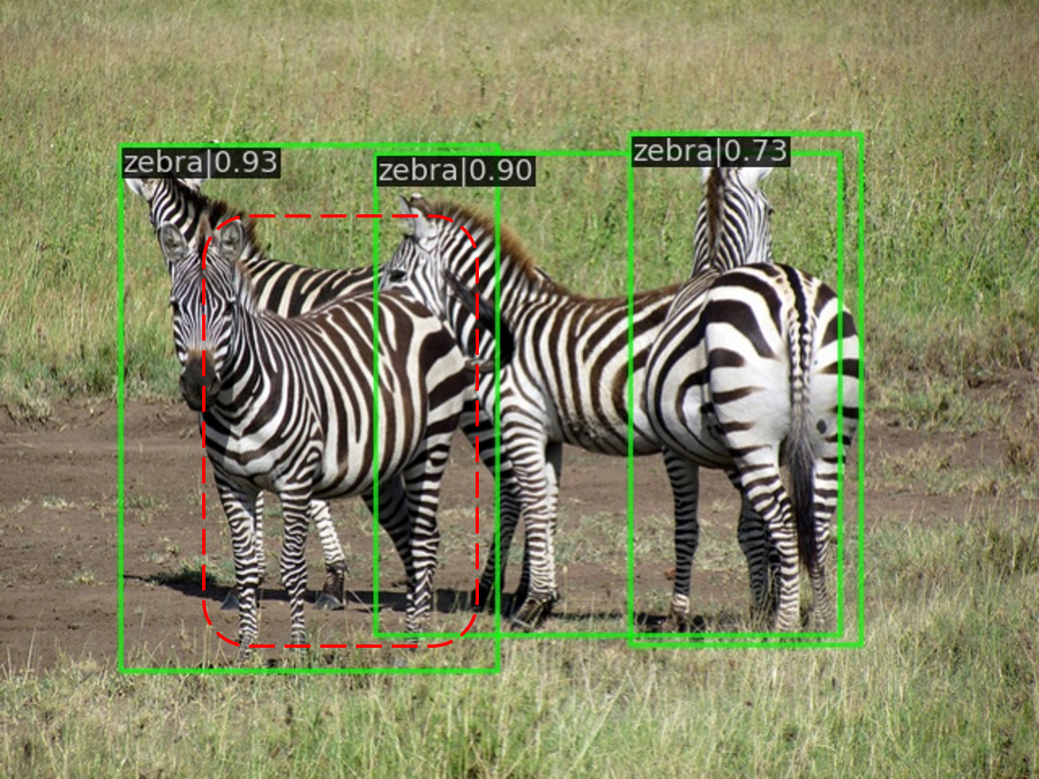}
  }
  \subfigure[Vanilla Student Model] {
    \label{fig: Detection Results of Vanilla Student Model}
    \includegraphics[width=0.20\linewidth]{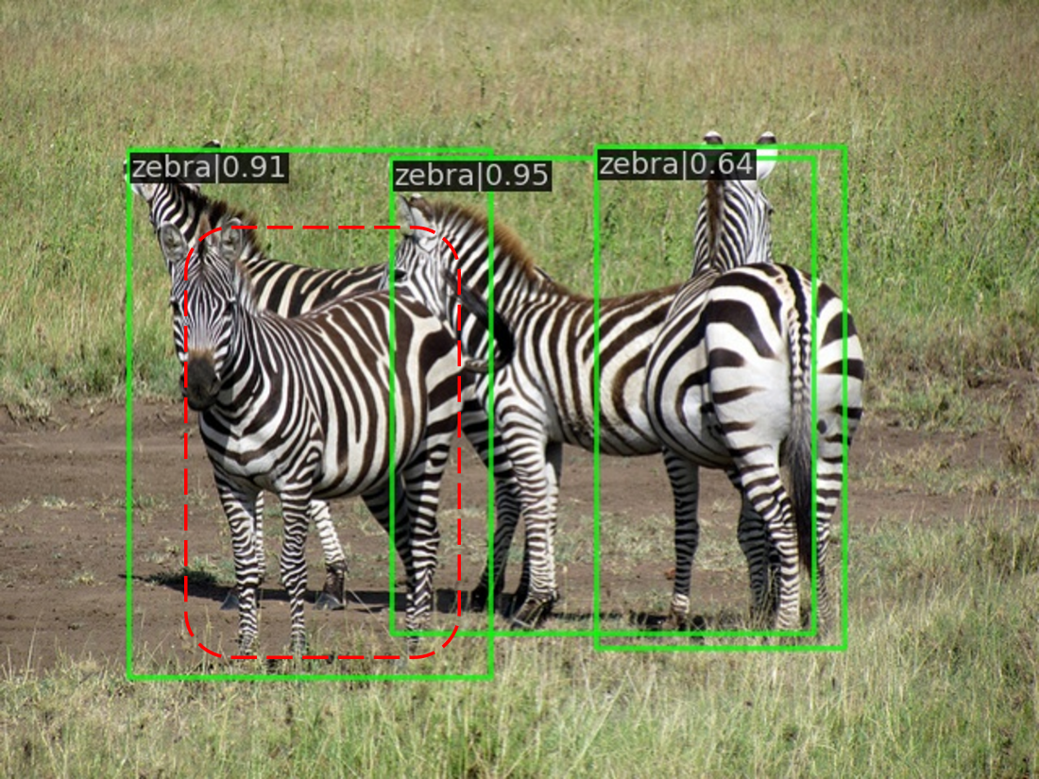}
  }
  \caption{Visualizations of detection results of different models. Red dashed bounding boxes indicate missing predictions. }
  \label{figure: Detection Results}
\end{figure*}

\begin{figure*}
  \centering
  \subfigure[Teacher Model] {
    \label{fig: Feature Map of Teacher Model}
    \includegraphics[width=0.20\linewidth]{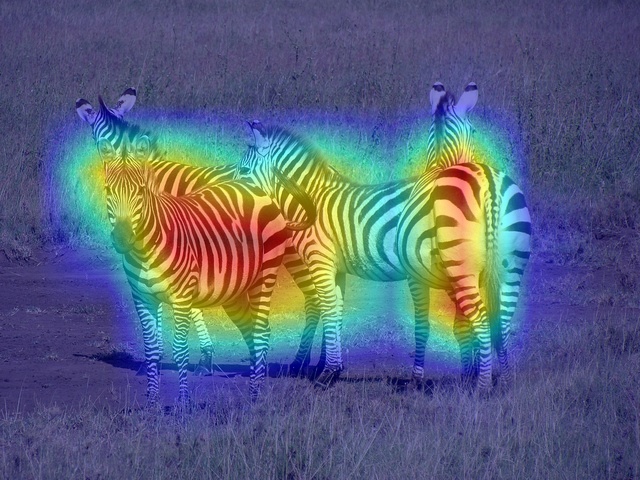}
  }
  \subfigure[Ours (+FGD)] {
    \label{fig: Feature Map of DFA}
    \includegraphics[width=0.20\linewidth]{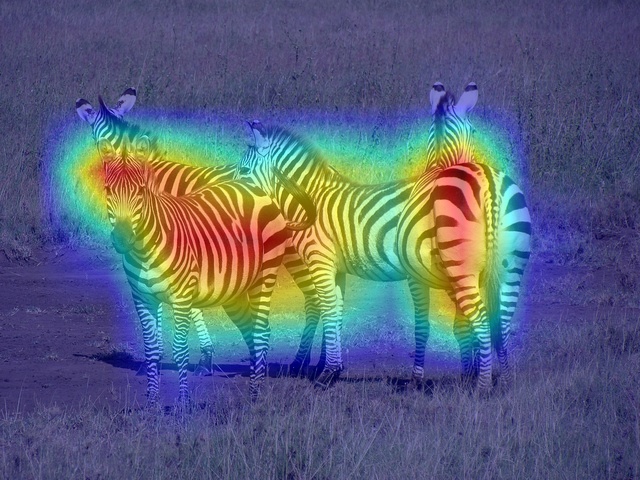}
  }
  \subfigure[FGD] {
    \label{fig: Feature Map of FGD}
    \includegraphics[width=0.20\linewidth]{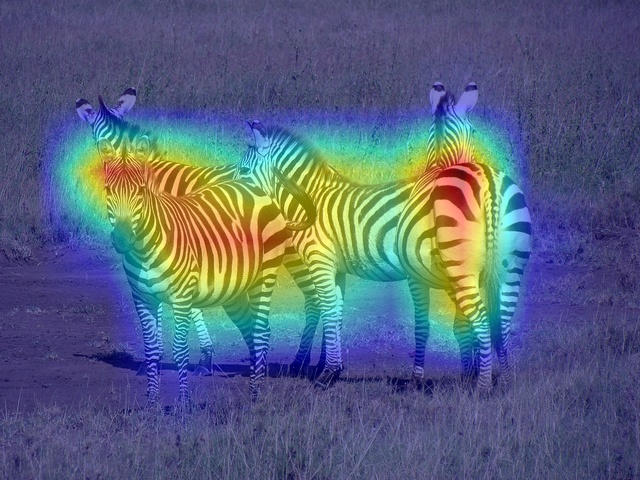}
  }
  \subfigure[Vanilla Student Model] {
    \label{fig: Feature Map of Vanilla Student Model}
    \includegraphics[width=0.20\linewidth]{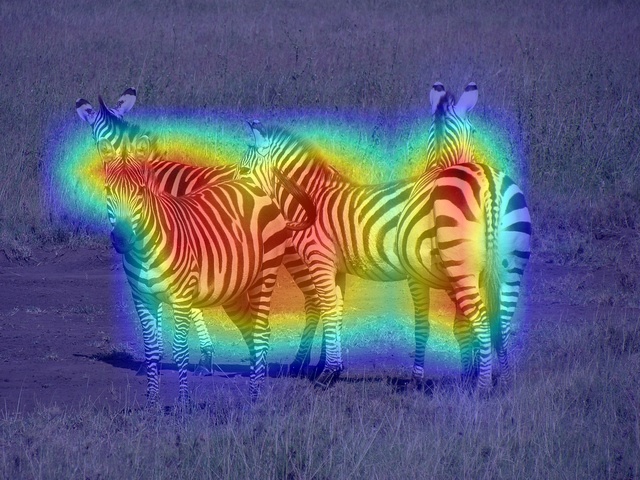}
  }
  \caption{Visualizations of feature maps of different models. Our method could enhance the matching between the feature maps of the student model and the teacher model as compared to the FGD baseline.}
  \label{figure: Feature Maps}
\end{figure*}

\begin{figure}
  \centering
    \subfigure[Confusion Matrix of Ours (+FGD)] {
    \label{fig: Confusion Matrix of DFA}
    \includegraphics[width=0.45\linewidth]{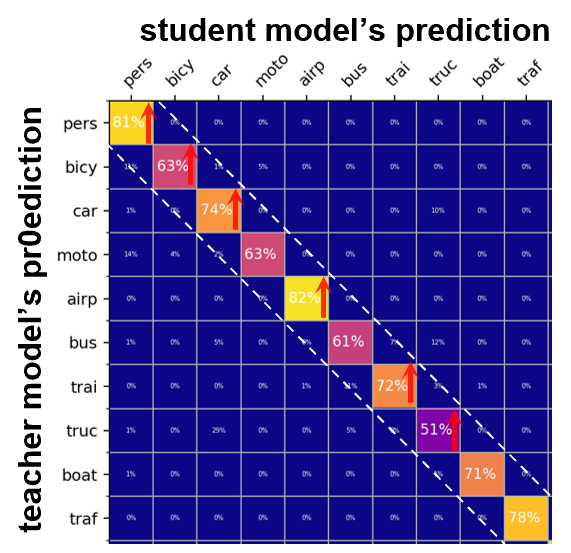}
  }
  \subfigure[Confusion Matrix of FGD] {
    \label{fig: Confusion Matrix of FGD}
    \includegraphics[width=0.468\linewidth]{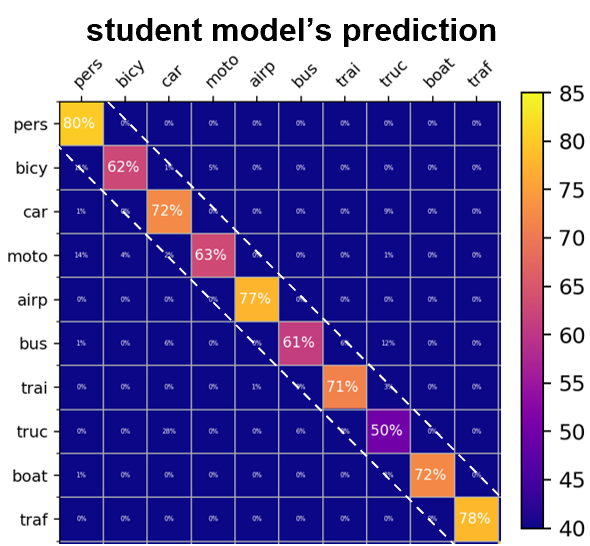}
  }

  \caption{Confusion matrix between the teacher model's predictions and the student model's predictions.}
  \label{figure: Confusion Matrix}
\end{figure}

\begin{table}
\small
    \centering
        \begin{tabular}{c|c|cccc}
        \toprule        
        Index &  method &  mAP & mAP$_S$ &   mAP$_M$  & mAP$_L$ \\
        \midrule
        1  & Teacher & 100 & 100 & 100 & 100 \\
        2  & Student & 52.7 & 35.3 & 51.0 & 59.4   \\
        3  & MGD~\cite{yang2022masked} &65.9 & 49.1 & 65.8 & 71.5 \\
        4  & FGD~\cite{yang2022focal} & 65.9 & 48.1 & 64.8 & 72.1 \\
        5  & Ours (+MGD) & \textbf{67.9}  & 52.2 & 67.5 & 73.2 \\
        5  & Ours (+FGD) & \textbf{68.1} & 52.3 & 67.2 & 74.1 \\
        \bottomrule
        \end{tabular}
    \caption{Results evaluated using the teacher model's predictions as ground truth. With higher mAP, the student model makes more identical predictions with the teacher model. }
    \label{Results with pseudo labels}
\end{table}

\subsection{Discussion and Analysis}

\textbf{More Knowledge Transferred from Teacher Model with Our Method}
We propose a data augmentation framework to mine inconsistent knowledge from the teacher model. To validate that our method can effectively transfer more knowledge from the teacher model, we evaluate the fidelity~\cite{jagielski2020high} of different KD methods, which measures the agreement between the output the teacher and student model on the same input. Firstly, we use the teacher model to make predictions on test set and only save predictions with confidence over 0.3 as pseudo labels. We then evaluate the mAP of the student model with the pseudo labels as the ground truth. Our experiments are conducted on RetinaNet object detector. The results presented in Table~\ref{Results with pseudo labels} demonstrate that knowledge distillation yields higher fidelity compared to training from scratch. Furthermore, our method shows an additional improvement in fidelity. On one hand, the results suggest that knowledge from the teacher model has not been fully distilled. On the other hand, the results also validate that our method can transfer more knowledge. 

We also visualize the detection results and deep feature maps of different models, as shown in Figure~\ref{figure: Detection Results} and Figure~\ref{figure: Feature Maps} respectively. From the detection results, we observe that our data augmentation method results in more accurate and consistent predictions with the teacher model. The results in Figure~\ref{figure: Feature Maps} show that the feature map of our method achieves the highest similarity with the teacher model. Furthermore, we compute confusion matrix of our method and make a comparison with the baseline, \ie, FGD. The results are presented in Figure~\ref{figure: Confusion Matrix}, where the values on the diagonal of the confusion matrix represent the ratio of predictions that match the teacher model's predictions. Our method achieves higher ratio of matching in most cases, which further validates that our method could transfer more knowledge.

\textbf{Generalizability to Other Scenarios}
To demonstrate the generalizability of our method, we evaluate its performance on a different task. Knowledge distillation is also used in backdoor defense~\cite{wu2022backdoorbench}, such as Neural Attention Distillation (NAD)~\cite{li2021neural}. In this experiment, we apply the Badnet~\cite{gu2017badnets} backdoor attack to a multimodal contrastive model, Declip~\cite{li2021supervision}, to obtain the victim model. Then, following the procedure of NAD, we use the model fine-tuned on clean data as the teacher model and purify the victim model using the distillation technique. Analogously, we apply our method with NAD and make a comparison. We evaluate the attack success rate (ASR) on the ImageNet-1K~\cite{deng2009imagenet} validation dataset using the zero-shot top-5 accuracy metric. The results in Table~\ref{Results on backdoor defense} demonstrate that our method can improve the existing NAD backdoor defense method by 33$\%$. \footnote{More backdoor attack and defense results are available in our codebase.}

\begin{table}[t]
\small
    \centering
       
        \begin{tabular}{c|c|cc}
            \toprule        
            Index &  method &  ASR & ASR Drop    \\
            \midrule
            1  & Victim & 96.7 & -   \\
            2  & NAD~\cite{li2021neural} & 82.88 & 13.82 \\
            3  & Ours (+NAD) &78.26 & 18.44 $(\uparrow 33 \%)$ \\
            \bottomrule
        \end{tabular}
    \caption{Results on backdoor defense task. The first row indicates the result of the victim model without defense.}
    \label{Results on backdoor defense}
\end{table}

\section{Conclusion}

In this paper, we reveal that current KD methods overlook inconsistent knowledge and lead to insufficient distillation from the teacher model. To address this issue, we introduce two available inconsistent knowledge for distillation, one related to frequency and the other associated with non-robust features. To distill these two inconsistent knowledge, we propose a sample-specific data augmentation that enhances distinct frequency components adaptively and an adversarial feature augmentation that mimics adversarial features. Furthermore, our method can be easily integrated into current KD methods for object detection and improve their performance.

\section*{Acknowledgement} 
This work was supported in part by the National Key R\&D Program of China (Grant No. 2022ZD0118100), in part by the National Natural Science Foundation of China (No.62025604, 62132006, 62261160653, 62006217, 62206009). 

\clearpage

\bibliographystyle{ACM-Reference-Format}
\bibliography{main}

\clearpage
\appendix
\counterwithin{table}{section}
\counterwithin{figure}{section}

\section*{Appendix}

\section{Sensitivity Study of $\beta$} 
In this paper, we incorporate inconsistent knowledge distillation with consistent knowledge distillation. In practice, the balance between them is important. For example, the hyper-parameter $\beta$ is introduced to weight the distillation of inconsistent knowledge concerning non-robust features. Here, we further conduct experiments to evaluate the effect of $\beta$. Specifically, we experiment on Mask RCNN with FGD. The results presented in Table~\ref{Sensitivity Study of bata} demonstrate that it is crucial to set $\beta$ to an appropriate value If $\beta$ is set too high, the model may become overly biased towards distilling inconsistent knowledge, whereas a low value of $\beta$ may lead to insufficient distillation of inconsistent knowledge. 

\begin{table}[h]
    \centering
        \begin{tabular}{c|cccc}
            \toprule        
           $\beta (\times 10^{-7})$    &   0.625 	&1.25 	&2.5 	&5.0\\
            \midrule
 
            mAP 	&42.2 	&42.2 	&\textbf{42.4} 	&42.1\\
            \bottomrule
        \end{tabular}
    \caption{Results of different $\beta$.}
    \label{Sensitivity Study of bata}
\end{table}

\section{Training Time Evaluation}
In this paper, we propose adversarial feature augmentation that utilizes adversarial examples as input to distill adversarial features, which incurs additional training overhead. To evaluate the training performance of our method, we benchmark our method on RetinaNet with FGD using 8 GPUs. Firstly, we generate adversarial examples in an offline manner, which takes approximately 3 hours. The training with additional data takes around 27 hours, while the baseline (vanilla training) takes about 24 hours.

In summary, (1) training with additional data results in an extra time cost of 3 hours (1/8 of the vanilla training), and (2) the total additional time cost is 6 hours when considering the time cost of offline data generation (1/6 of the vanilla training).

In fact, since the generated adversarial examples can be reused, there is no need to regenerate them every time before training. Nonetheless, improving computational efficiency remains an area of future study.

\end{document}